# Solution Subset Selection for Final Decision Making in Evolutionary Multi-Objective Optimization

**Hisao Ishibuchi**[1], **Lie Meng Pang** and **Ke Shang**
University Key Laboratory of Evolving Intelligent Systems of Guangdong Province
Department of Computer Science and Engineering, Southern University of Science and Technology
hisao@sustech.edu.cn, panglm@sustech.edu.cn, kshang@foxmail.com

## Abstract

In general, a multi-objective optimization problem does not have a single optimal solution but a set of Pareto optimal solutions, which forms the Pareto front in the objective space. Various evolutionary algorithms have been proposed to approximate the Pareto front using a pre-specified number of solutions. Hundreds of solutions are obtained by their single run. The selection of a single final solution from the obtained solutions is assumed to be done by a human decision maker. However, in many cases, the decision maker does not want to examine hundreds of solutions. Thus, it is needed to select a small subset of the obtained solutions. In this paper, we discuss subset selection from a viewpoint of the final decision making. First we briefly explain existing subset selection studies. Next we formulate an expected loss function for subset selection. We also show that the formulated function is the same as the IGD plus indicator. Then we report experimental results where the proposed approach is compared with other indicator-based subset selection methods.

## 1 Introduction

An $m$-objective minimization problem is written as

$$\text{Minimize } f(x) = (f_1(x), f_2(x), ..., f_m(x)), \quad (1)$$
$$\text{subject to } x \in X, \quad (2)$$

where $f_i(x)$ is the $i$th objective to be minimized ($i = 1, 2, ..., m$), $x$ is a decision vector, and $X$ is the feasible region of $x$. In general, some or all objectives in (1) are conflicting with each other. Thus, the multi-objective problem in (1)-(2) has no single absolute optimal solution that simultaneously optimizes all objectives. Instead, it has a set of solutions that are non-dominated with each other.

Feasible solutions of the multi-objective problem in (1)-(2) are compared using the following Pareto dominance relation: When $f_i(x^A) \leq f_i(x^B)$ holds for all $i$ and $f_j(x^A) < f_j(x^B)$ holds for at least one $j$, we say that solution $x^A$ dominates solution $x^B$ (i.e., $x^A$ is better than $x^B$). If solution $x^B$ is not dominated by any other solutions in $X$, $x^B$ is called a Pareto optimal solution. The set of all Pareto optimal solutions is the Pareto optimal solution set. The projection of the Pareto optimal solution set to the objective space is the Pareto front. A solution set is referred to as a non-dominated solution set when all solutions in the solution set are non-dominated.

Using the Pareto optimal solution set $PS$, the ideal point $z^{\text{Ideal}} = (z_1^{\text{Ideal}}, z_2^{\text{Ideal}}, ..., z_m^{\text{Ideal}})$ and the nadir point $z^{\text{Nadir}} = (z_1^{\text{Nadir}}, z_2^{\text{Nadir}}, ..., z_m^{\text{Nadir}})$ are defined in the objective space as

$$z_i^{\text{Ideal}} = \min\{f_i(x) | x \in PS\}, i = 1, 2, ..., m, \quad (3)$$
$$z_i^{\text{Nadir}} = \max\{f_i(x) | x \in PS\}, i = 1, 2, ..., m. \quad (4)$$

It should be noted that $z_i^{\text{Ideal}}$ is the best objective value of the $i$th objective in the feasible region $X$ while $z_i^{\text{Nadir}}$ is not the worst objective value in $X$:

$$z_i^{\text{Ideal}} = \min\{f_i(x) | x \in X\}, i = 1, 2, ..., m, \quad (5)$$
$$z_i^{\text{Nadir}} \neq \max\{f_i(x) | x \in X\}, i = 1, 2, ..., m. \quad (6)$$

In the last three decades, a large number of evolutionary multi-objective optimization (EMO) algorithms have been proposed to find well-distributed solutions over the entire Pareto front [Deb, 2001]. Hundreds of solutions are obtained by a single run of an EMO algorithm. If an EMO algorithm has an unbounded external archive [Ishibuchi *et al.*, 2020], tens of thousands of solutions can be obtained. For example, in [Ishibuchi *et al.*, 2014], 220,298 non-dominated solutions were obtained by MOEA/D [Zhang and Li, 2007] for a four-objective 500-item knapsack problem.

However, in many cases, human decision makers do not want to examine such a large number of solutions when they need to choose a single final solution [Preuss and Wessing, 2013]. Human decision makers also have biological limitation about the amount of information they can handle simultaneously, which is known as "the magical number seven plus or minus two" [Miller, 1956]. Thus, a post-processing procedure (which is a prescreening procedure for the final decision making) is needed to select only a small number of solutions. In [Preuss and Wessing, 2013], a post-processing procedure was proposed to select five solutions from all the examined solutions during the execution of an evolutionary algorithm on a multi-modal optimization problem (i.e., a single-objective problem with a number of different optimal solutions). Our focus in this paper is to select a small number of promising solutions from a large number of obtained solutions for a multi-objective optimization problem. We assume that a single final solution is chosen by the decision maker from the selected solution subset.

---
[1] Corresponding Author.

In the EMO community, solution subset selection based on the hypervolume indicator [Zitzler and Thiele, 1998] has been actively studied under the name of hypervolume subset selection (HSS [Kuhn et al., 2016]). The HSS problem is to find a pre-specified number of solutions from a given solution set to maximize their hypervolume. A number of exact optimization algorithms as well as heuristic greedy algorithms have been proposed for the HSS problem. However, their use for the final decision making has several difficulties. The main difficulty is that the relation between the hypervolume maximization and the final decision making is not clear. The hypervolume maximization does not necessarily mean the selection of promising solutions for the final decision making. Another difficulty is huge computation load in the case of a large solution set with many objectives.

In this paper, we discuss solution subset selection from a viewpoint of the final decision making. Let us assume that we have a solution set $S$ obtained by an EMO algorithm. Our problem is to select a subset $A$ with a pre-specified number (say, $k$) of solutions (i.e., $|A| = k$) from $S$. We assume that a single final solution is chosen from the subset $A$. We also assume that an integer value for $k$ is given by the decision maker (e.g., $k = 10$), which is the number of solutions that he/she wants to examine as candidates for a single final solution. In order to select $A$ for the final decision making, we formulate an expected loss function. The loss of choosing $a$ from $A$ instead of $s$ from $S$ as the single final solution (when $s$ is not included in $A$) is defined by the amount of the deterioration in the objective space caused by changing the choice from $s$ to $a$. If $s$ is included in $A$ (i.e., $a = s$), the loss is zero. The expected loss is calculated as the average loss over all solutions in $S$. We propose to use the formulated function in order to minimize the expected loss caused by subset selection. Our idea is to present the selected subset $A$ with the minimum expected loss to the decision maker together with the $m$ extreme solutions with the best value for each objective. The $m$ extreme solutions may be used by the decision maker to know the range of the obtained solution set $S$ in the objective space. The single final solution will be chosen from the selected subset $A$ (not from the extreme solutions since they are not well-balanced tradeoff solutions, i.e., since they usually have very bad objective values for some objectives).

This paper is organized as follows. In Section 2, we briefly explain existing HSS studies. We also point out some difficulties in their use to select promising solutions for the final decision making. In Section 3, we formulate an expected loss function for subset selection. We also show that the formulated function is the same as the inverted generational distance plus (IGD$^+$) indicator [Ishibuchi et al., 2015]. In Section 4, we report subset selection results on some test problems by the three indicators: Hypervolume, inverted generational distance (IGD [Coello and Sierra, 2004]) and IGD$^+$. Finally, we conclude this paper in Section 5.

## 2 Hypervolume Subset Selection

### 2.1 Hypervolume Subset Selection Problem

The hypervolume (HV) of a point $s$ in the $m$-dimensional objective space is the area ($m=2$), volume ($m=3$) or hypervolume ($m>3$) of the region dominated by $s$ and bounded by a reference point $r$:

$$HV(s, r) = Volume(\{z \mid s \le z \le r\}), \qquad (7)$$

where $Volume(.)$ calculates the area, volume or hypervolume.

The hypervolume of a point set $S = \{s_1, s_2, ..., s_n\}$ is calculated for the reference point $r$ in the same manner as (7):

$$HV(S, r) = Volume\left(\bigcup_{s_i \in S}\{z \mid s_i \le z \le r\}\right). \qquad (8)$$

The hypervolume contribution (HVC) of a point $s_i$ in $S = \{s_1, s_2, ..., s_n\}$ is defined for the reference point $r$ as

$$HVC(s_i, S, r) = HV(S, r) - HV(S - \{s_i\}, r). \qquad (9)$$

The hypervolume subset selection (HSS) problem can be written for a given point set $S$ and a given integer $k$ ($k \in \{1, 2, ..., |S|\}$) as follows [Kuhn et al., 2016]:

**Hypervolume Subset Selection (HSS) Problem:**
Find a subset $A$ of $S$ such that

$$|A|=k \text{ and } HV(A)=\max\{HV(B) \mid B \subset S, |B|=k\}. \qquad (10)$$

It should be noted that other performance indicators can be used in (10) for subset selection. For example, we can formulate the IGD subset selection problem and the IGD$^+$ subset selection problem in the same manner.

### 2.2 Exact Optimization Approaches

In principle, the optimal solution of the HSS problem in (10) can be obtained by examining all combinations of $k$ points selected from the given $n$ points in $S$ (i.e., by calculating the hypervolume of each of the $_nC_k$ subsets with $k$ points from $S$).

Since the total number $_nC_k$ of subsets of size $k$ severely increases with $n$, the use of a straightforward enumeration method is impractical for a large point set $S$. A number of exact optimization algorithms have been proposed to solve the HSS problem efficiently. Most of them were proposed for the HSS problem in low-dimensional objective spaces. For example, [Bringmann et al., 2014a] proposed an exact algorithm for the two-dimensional HSS problem (i.e., $m=2$). Using the proposed algorithm, the optimal subset of all the examined solutions was selected and compared with the final population for some EMO algorithms in [Bringmann et al., 2014b]. A similar exact algorithm was also proposed in [Kuhn et al., 2016] for the case of $m=2$.

A general exact algorithm, which is applicable to the case of $m>2$, was proposed in [Bringmann and Friedrich, 2010]. However, in their computational experiments, the size of the point set $S$ was very small ($n = 20$ for $m = 5$, and $n = 10$ for $m = 6$ and $m = 8$). Two general exact algorithms were proposed in [Gomes et al., 2018]. However, they did not show any experimental results for high-dimensional cases with $m > 4$. Moreover, the size of the point set was small ($n = 50$ for $m = 4$). [Groz and Maniu, 2019] proposed an efficient algorithm for a special case where $k$ or $(n - k)$ is very small. In their computational experiments, $k$ was specified as $k = 2$ for the four-dimensional HSS problem ($m = 4$). They did not examine their algorithm for the case of $m > 4$.

The applications of the above-mentioned exact algorithms are limited to the two- and three-objective problems except for some special cases such as $k = 2$ and $n = 10, 20, 50$.

### 2.3 Incremental Greedy Approaches

For the HSS problem, incremental and decremental greedy algorithms were proposed. Since our subset selection problem is to find a small number of promising candidate solutions, incremental algorithms are more efficient than decremental ones. An incremental greedy algorithm for the HSS problem has the following general framework. First the best single solution with the largest hypervolume value is selected. Next a single solution with the largest hypervolume contribution to the selected solution set is found and added. This step is iterated to select a pre-specified number of solutions. In this framework, the main computation load is hypervolume contribution calculation. [Guerreiro et al., 2016] proposed an efficient calculation method and evaluated its efficiency on two- and three-objective problems.

One disadvantage of the incremental greedy algorithm is that the selected subset can have a large difference from the optimal one with the maximum hypervolume, especially when the number of selected solutions is small. To visually explain this issue, we show the greedy incremental subset selection results in Fig. 1 and the corresponding optimal subsets for hypervolume maximization in Fig. 2.

In these figures, the Pareto front is the blue straight line. The reference point $r$ is specified as $r = (2, 2)$. Since the optimal distribution of solutions for hypervolume maximization can be theoretically calculated for the two-objective linear Pareto front [Auger et al, 2012], we depicted Fig. 1 by assuming that an infinitely large number of points are given over the Pareto front. From Fig. 1 and Fig. 2, we can see that the selected subsets by the incremental greedy algorithm are clearly different from the optimal ones when $k = 2, 4, 6$.

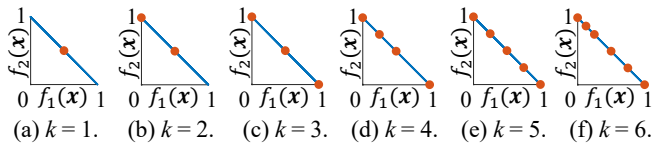

(a) $k = 1$. (b) $k = 2$. (c) $k = 3$. (d) $k = 4$. (e) $k = 5$. (f) $k = 6$.

Figure 1. Results of the HV-based incremental greedy method.

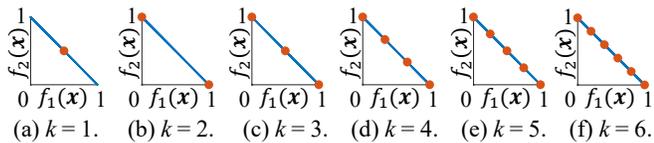

(a) $k = 1$. (b) $k = 2$. (c) $k = 3$. (d) $k = 4$. (e) $k = 5$. (f) $k = 6$.

Figure 2. Optimal distributions of solutions for HV maximization.

### 2.4 Metaheuristic Approaches

When the use of exact optimization algorithms is impractical due to their huge computation load, it may be worth trying some metaheuristic algorithms. In [Ishibuchi et al., 2014], a genetic algorithm (GA) was used for the HSS problem. In their GA-based approach, a subset $A$ of the solution set $S$ with $n$ solutions was coded by a binary string $\boldsymbol{b}$ of the length $n$ as

$$\boldsymbol{b} = b_1 b_2 ... b_n, \quad (11)$$
$$A = \{s_i \mid b_i = 1, i = 1, 2, ..., n\}. \quad (12)$$

In our computational experiments, we use the following implementation of a genetic algorithm for the HSS problem:

- Initialization: Randomly generated population of $\mu$ binary strings ($\mu = 100$ in our computational experiments). All strings are different, and the number of 1 in each string is $k$.
- Parent selection: Binary tournament selection.
- Crossover: One-point crossover.
- Crossover probability: 1.0.
- Mutation: Bit-flip mutation with biased probability.
- Mutation probability: $1/k$ for $1 \rightarrow 0$, $1/(n-k)$ for $0 \rightarrow 1$.
- Repair: A repair procedure is used so that all offspring strings have exactly $k$ bits with 1. If the number of 1 in an offspring string is not $k$, it is repaired by choosing the required number of bits randomly and changing their values from 1 to 0 or from 0 to 1.
- The number of offspring in each generation: $\mu$.
- Generation update: $(\mu + \mu)$-ES style update where the best $\mu$ strings are selected as the next generation from the current $\mu$ strings and the generated $\mu$ offspring strings. All duplicated strings are removed during the generation update phase.
- Termination: After a pre-specified number of generations (2000 generations in our computational experiments).

The biased mutation operator is used since the number of selected solutions (i.e., the number of 1) is much smaller than the total number of solutions in $S$ (i.e., the string length $n$). The repair procedure is used so that each string includes exactly $k$ solutions. All duplicated strings are removed during the generation update phase in order to maintain the diversity of strings in each generation.

Good subsets are obtained by the GA-based method within a reasonable computation time. However, there is no theoretical guarantee about the quality of the obtained subsets. This is the main difficulty of the GA-based method.

### 2.5 Difficulties of HV-based Subset Selection

In addition to the above-mentioned difficulties, HV-based subset selection has the following two difficulties in its use for the final decision making. One is that there is no clear relation between the hypervolume maximization and the final decision making. When only a small number of solutions are presented, it is very likely that the decision maker wants to know a convincing reason for their selection. We have no convincing reason why the hypervolume should be used in subset selection for the final decision making.

The other difficulty is the strong dependency of subset selection results on the reference point specification. To demonstrate this difficulty, we first applied NSGA-II [Deb et

*al.*, 2002a] with the population size 500 to the normalized three-objective minus-DTLZ1 problem [Ishibuchi et al., 2017]. This problem has the linear Pareto front specified by $f_1+f_2+f_3 = 2$ and $0 \leq f_i \leq 1$ for $i = 1, 2, 3$. By a single run of NSGA-II, 500 non-dominated solutions were obtained (i.e., $n = 500$). Then our GA-based approach was applied to the obtained solution set for selecting nine solutions (i.e., $k = 9$). We examined three settings of the reference point $r = (r, r, r)$: $r = 1.0, 1.2, 1.5$. It should be noted that the ideal point and the nadir point of the normalized minus-DTLZ1 problem are (0, 0, 0) and (1, 1, 1), respectively. Experimental results are shown in Fig. 3, which is the median HV result for each setting over 31 runs. Blue and red points are the obtained 500 solutions and the selected 9 solutions, respectively.

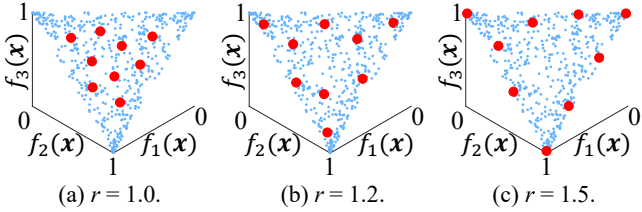

(a) $r = 1.0$.  (b) $r = 1.2$.  (c) $r = 1.5$.

Figure 3. Results of the HV-based subset selection.

Fig. 3 clearly shows the strong dependency of the selected subsets on the reference point specification. When the nadir point (1, 1, 1) was used as the reference point in Fig. 3 (a), all the selected solutions are around the center of the Pareto front. By increasing the value of $r$ (i.e., by increasing the distance from the reference point to the Pareto front), more solutions were selected near the edges of the Pareto front. There is no clear reason for the use of any specific reference point in subset selection for the final decision making whereas the subset selection results strongly depend on its setting as shown in Fig. 3. This is the main difficulty of the HV-based subset selection when it is used for the final decision making. We cannot explain to the decision maker why a specific reference point is used in subset selection.

## 3 Expected Loss Function

In this section, we formulate an expected loss function to give a clear meaning to subsect selection for the final decision making. As in the previous section, we assume that we have a set of $n$ solutions (i.e., $n$ points) in the $m$-dimensional objective space: $S = \{s_1, s_2, ..., s_n\}$. Our problem is to select a subset $A$ of size $k$ from $S$: $A = \{a_1, a_2, ..., a_k\} \subset S$. The selected subset $A$ is presented to the decision maker. A single final solution is selected from $A$ by the decision maker.

Let $s$ be the (unknown) preferred solution by the decision maker in the solution set $S$. If $s$ is in the selected solution subset $A$, it will be selected as the final solution by the decision maker. In this case, we have no loss by the selection of $A$ from $S$. Since the number of selected solutions in $A$ is usually much smaller than the total number of solutions in $S$ (i.e., $|A| \ll |S|$), it is likely that $s$ is not included in $A$. In this case, we assume that the nearest solution $a$ in $A$ to $s$ will be selected as the final solution by the decision maker. To formulate the loss by the selection of $A$ in this context, we need to discuss the following two issues: the choice of the nearest solution $a$ and the definition of the loss by choosing $a$ instead of $s$.

First we discuss the second issue. In Fig. 4, we show three cases of the locations of $s = (s_1, s_2)$ and $a = (a_1, a_2)$ in the objective space of a two-objective minimization problem. In Fig. 4 (a), $a$ is dominated by $s$. In this case, we use the distance between $a$ and $s$ as the loss by choosing $a$ instead of $s$ (actually this case does not happen since all solutions in $S$ are non-dominated). In Fig. 4 (b), $a$ and $s$ are non-dominated. Since $a$ is better than $s$ for $f_2(x)$, their distance with respect to $f_2(x)$ is not the loss. The loss is their distance with respect to $f_1(x)$ in Fig. 4 (b). On the contrary, in Fig. 4 (c), $a$ is better than $s$ for $f_1(x)$. Thus, the loss is their distance only for $f_2(x)$.

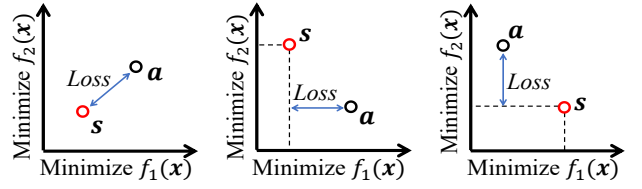

(a) $s_1 < a_1$ and $s_2 < a_2$.  (b) $s_1 < a_1$ and $s_2 > a_2$.  (c) $s_1 > a_1$ and $s_2 < a_2$.

Figure 4. Three cases of the locations of $s$ and $a$.

From these discussions, we define the loss by choosing $a = (a_1, a_2)$ instead of $s = (s_1, s_2)$ for the two-objective minimization problem as follows:

$$Loss(a, s) = \sqrt{(\max\{0, a_1 - s_1\})^2 + (\max\{0, a_2 - s_2\})^2} . \quad (13)$$

This formulation is generalized to the case of $m$ objectives:

$$Loss(a, s) = \sqrt{\sum_{i=1}^{m}(\max\{0, a_i - s_i\})^2} . \quad (14)$$

In this formulation, the distance for the $i$th objective with $a_i < s_i$ is ignored since $a$ is better than $s$ with respect to the $i$th objective. Only those objectives with $a_i > s_i$ are used in the calculation of the loss function in (14). In the case of maximization problems, $a_i - s_i$ is replaced with $s_i - a_i$ in (14).

Next we discuss the choice of a single final solution from the subset $A$ by the decision maker whose preferred solution among $S$ is $s$. It may be reasonable to assume that the decision maker will choose the solution with the minimum loss from $A$. Based on this assumption, we define the loss by selecting the subset $A$ from $S$ as follows:

$$Loss(A, s) = \min_{a \in A}\{Loss(a, s)\} . \quad (15)$$

This formulation defines the loss by the selection of the solution subset $A$ for the decision maker whose preferred solution is $s$. However, we have no information about the preferred solution $s$. Thus we assume that each solution $s$ in $S$ has the same probability to be selected as a single final solution. Based on this assumption, we define the expected loss by the selection of the solution subset $A$ from $S$ as

$$Loss(A, S) = \frac{1}{|S|}\sum_{s \in S}\min_{a \in A}\{Loss(a, s)\}. \quad (16)$$

Our proposal is to select the solution subset that minimizes the expected loss. Our problem is formulated as follows:

**Minimum Loss Subset Selection Problem:**
Find a subset $A$ of $S$ such that
$$|A|=k \text{ and } Loss(A, S)=\min\{Loss(B, S)|B\subset S, |B|=k\}. \quad (17)$$

Using (14), (16) is rewritten as follows:
$$Loss(A, S) = \frac{1}{|S|} \sum_{s \in S} \min_{a \in A} \left\{ \sqrt{\sum_{i=1}^{m}(\max\{0, a_i - s_i\})^2} \right\}, \quad (18)$$

which is the same as the IGD$^+$ indicator [Ishibuchi *et al.*, 2015] of the solution subset $A$ for the reference point set $S$. Thus, our subset selection problem can also be written as:

**IGD$^+$ Subset Selection Problem:**
Find a subset $A$ of $S$ such that
$$|A|=k \text{ and } IGD^+(A, S)=\min\{IGD^+(B, S)|B\subset S, |B|=k\}, \quad (19)$$
where
$$IGD^+(A, S) = \frac{1}{|S|} \sum_{s \in S} \min_{a \in A} \left\{ \sqrt{\sum_{i=1}^{m}(\max\{0, a_i - s_i\})^2} \right\}. \quad (20)$$

The IGD$^+$ indicator is a modified version of the IGD indicator. One advantage of IGD$^+$ over IGD is that IGD$^+$ is weakly Pareto compliant whereas IGD is not Pareto compliant (for further discussions about the Pareto compliance of performance indicators, see [Zitzler *et al.*, 2007]).

The IGD indicator of the solution subset $A$ with respect to the reference point set $S$ is written as
$$IGD(A, S) = \frac{1}{|S|} \sum_{s \in S} \min_{a \in A} \left\{ \sqrt{\sum_{i=1}^{m}(a_i - s_i)^2} \right\}. \quad (21)$$

IGD is the average distance from each solution $s$ in $S$ to the nearest solution $a$ in $A$. The dominance relation between $a$ and $s$ is not taken into account (i.e., the three cases in Fig. 4 are handled in the same manner). It should be noted that IGD and IGD$^+$ were originally proposed for performance comparison of EMO algorithms (not for subset selection).

Our proposal is to present $k$ solutions with the minimum expected loss (i.e., the minimum IGD$^+$ value) to the decision maker. If he/she wants to know the best achievable value of each objective, we can easily choose and show the $m$ extreme solutions with the best objective value for each objective.

## 4 Experimental Results

In this section, we examine subset selection based on the three indicators (hypervolume (HV), IGD, and the expected loss (IGD$^+$)) through computational experiments on various test problems: two-objective DTLZ2 [Deb et al., 2002b], two-objective minus-DTLZ2 [Ishibuchi *et al.*, 2017], two-objective wave-3 and wave-5 [Guerreiro *et al.*, 2016], and five-objective distance minimization [Köppen and Yoshida, 2007]. The reference point $r$ for HV calculation was specified by the formulation proposed for the fair performance comparison in [Ishibuchi et al., 2018] as (1.125, 1.125) for two-objective problems and (2, 2, 2, 2, 2) for five-objective problems. For subset selection, we used the genetic algorithm explained in Subsection 2.4. The number of solutions to be selected was specified as 9 (i.e., $k=9$).

A solution set $S$ for each test problem (except for the wave-3 and wave-5 problems) was generated by NSGA-II with the population size 500. In each wave problem, 500 points were randomly sampled from the Pareto front since no information except for the Pareto front (e.g., objective functions, decision variables) is available. The genetic algorithm for subset selection was applied to the generated solution set 31 times. Among them, a single run with the median indicator value is selected. The subset selection result by the selected run is shown in the normalized objective space with the ideal point (0, ..., 0) and the nadir point (1, ..., 1). The normalization is used only for visualization. NSGA-II and the genetic algorithm for subset selection were performed in the original objective space of each test problem.

### 4.1 Two-objective Concave and Convex Fronts

In Fig. 5 and Fig. 6, we show subset selection results for the two-objective DTLZ2 and minus-DTLZ2 problems, respectively. In each figure, the nine red points are the selected solutions. The blue curve shows the solution set $S$.

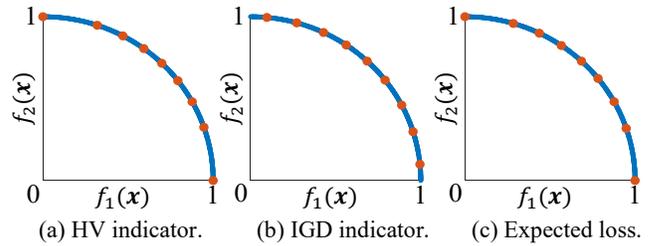

Figure 5. Results on the two-objective DTLZ2.

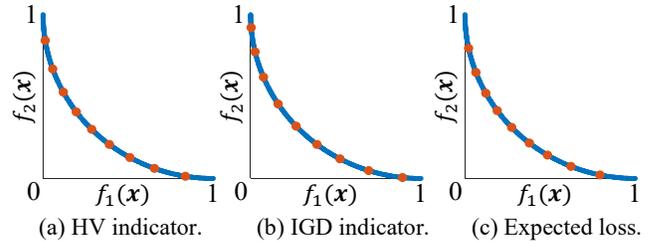

Figure 6. Results on the two-objective minus-DTLZ2.

When we used IGD, the uniformly distributed solutions were obtained independent of the shape of the Pareto front in Fig. 5 (b) and Fig. 6 (b). However, the obtained subsets by HV and the expected loss function depend on the shape of the Pareto front. In the case of the convex Pareto front in Fig. 6 (a) and (c), more solutions were obtained around the center of the Pareto font that is close to the ideal point (0, 0). This is consistent with our intuition that a single final solution will be selected around the center of the convex Pareto front in Fig. 6. In the case of the concave Pareto front in Fig. 5 (a) and (c), the two extreme solutions were selected. This is also consistent with our intuition that the two extreme solutions may have some chances to be selected as a single final solution (since the center region of the concave Pareto front is not close to the ideal point in Fig. 5). The obtained subsets by HV and the expected loss function are very similar in Fig. 5 and similar in Fig. 6. It should be noted that the HV-based subset selection results depend on the setting of the reference point.

## 4.2 Two-objective Wave Fronts

In Fig. 7 and Fig. 8, we show experimental results on the wave-3 and wave-5 test problems, respectively.

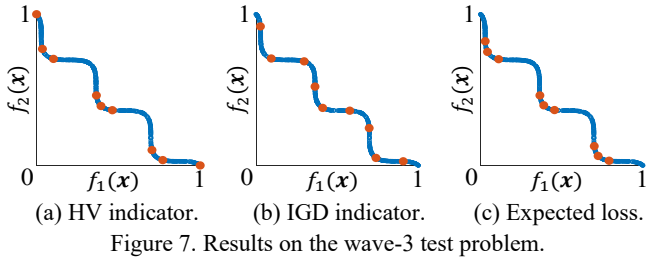

(a) HV indicator.    (b) IGD indicator.    (c) Expected loss.
Figure 7. Results on the wave-3 test problem.

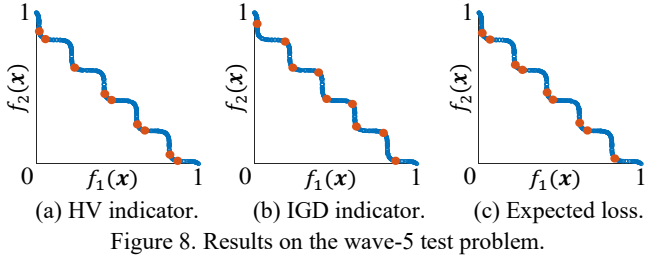

(a) HV indicator.    (b) IGD indicator.    (c) Expected loss.
Figure 8. Results on the wave-5 test problem.

When we used the expected loss function in Fig. 7 (c), all solutions were selected from the three convex regions. These convex regions (which are closer to the ideal point than the other regions) are called "knees". A single final solution is likely to be selected from those knee regions. In Fig. 7 (a) with the HV indicator, the two extreme solutions were also selected in addition to the seven knee solutions. These observations suggest that the knee regions are favored by the expected loss function more strongly than the HV indicator. This is also observed in Fig. 6. In Fig. 8 (a) and (c), all solutions were selected from the knee regions. In Fig. 7 (b) and Fig. 8 (b) with IGD, uniformly distributed solutions were obtained independent of the Pareto front shape.

## 4.3 Five-objective Distance Minimization Problem

An $m$-objective distance minimization problem is to minimize the distance to each of the given $m$ points. Here we used a five-objective version with five points (0, 1), (0.95, 0.31), (0.59, −0.81), (−0.59, −0.81) and (−0.95, 0.31) in the decision space $[−1, 1]^2$. The pentagonal region inside the five points are Pareto optimal. Experimental results are shown in Fig. 9 in the decision space where blue and red points are solutions in the solution set $S$ and the selected subset $A$, respectively. In Fig. 9, clearly different subsets were obtained by the three methods. However, it is difficult to say which is the best subset for the final decision making. In this situation, we can present Fig. 9 (c) to the decision maker since it has a clear meaning for the final decision making (i.e., the subset in Fig. 9 (c) has the minimum expected loss).

One difficulty in the proposed approach is the assumption that each solution in $S$ has the same selection probability as a single final solution. Due to this assumption, solutions in crowded regions are more likely to be selected to minimize the expected loss. As shown in Fig. 9 (c), no solutions are selected in less crowded regions (e.g., the center of the pentagon). This difficulty can be addressed in various manners (which are interesting future research topics). One idea is to assign a different weight to each point in $S$. Larger weights are assigned to points in less crowded regions. The expected loss is calculated as the weighted average loss. Another idea is to use the proposed approach several times to gradually decrease the number of solutions from $n$ to $k$ (e.g., from 500 to 100, then from 100 to 9). The selected solutions in the current iteration are used as the solution set $S$ in the next iteration. A different subset selection method can also be used in the first iteration. In Fig. 10 (a), we show 100 solutions selected by the distance-based selection method [Singh *et al.*, 2019]. Fig. 10 (b) and Fig. 10 (c) show the subset selection results from the 100 points in Fig. 10 (a). The uniformity of the selected solutions seems to be improved from Fig. 9 (b) to Fig. 10 (b) (and from Fig. 9 (c) to Fig. 10 (c)).

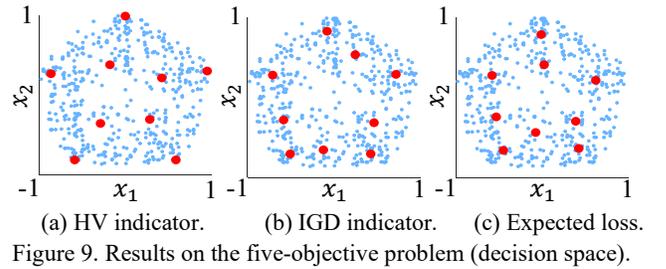

(a) HV indicator.    (b) IGD indicator.    (c) Expected loss.
Figure 9. Results on the five-objective problem (decision space).

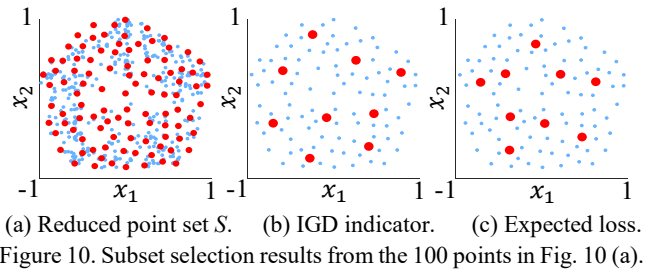

(a) Reduced point set $S$.    (b) IGD indicator.    (c) Expected loss.
Figure 10. Subset selection results from the 100 points in Fig. 10 (a).

## 5 Concluding Remarks

In this paper, we discussed solution subset selection for the final decision making. First we pointed out some difficulties in the use of hypervolume-based subset selection methods for the final decision making. Then we formulated an expected loss function, which gives a clear meaning to the selected solution subset in the context of the final decision making. We also showed that the formulated expected loss function is the same as the IGD$^+$ indicator when the given solution set is used as the reference point set. Finally we demonstrated that knee regions are favored by the expected loss function. This paper is the first attempt to try to bridge a gap through subset selection between a large number of solutions obtained by an EMO algorithm and the choice of a single final solution by the decision maker. Some future research topics have become clear by this attempt. One is the handling of a biased solution set as explained in Fig. 9 and Fig. 10. Development of efficient subset selection algorithms for IGD$^+$ is an important research topic since subset selection has been actively studied only for the hypervolume indicator (which is not applicable to a large solution set of a many-objective problem).